\documentclass{article}

\usepackage[preprint,nonatbib]{neurips_2026}

\usepackage[numbers, compress]{natbib} 
\usepackage[utf8]{inputenc}
\usepackage[T1]{fontenc}
\usepackage{amsmath,amssymb}
\usepackage{graphicx}
\usepackage{booktabs}
\usepackage{hyperref}
\usepackage{xcolor}
\usepackage{wrapfig}
\usepackage{siunitx}    
\usepackage[table]{xcolor} 
\usepackage{multirow}   
\usepackage[capitalise]{cleveref} 
\usepackage{caption}
\usepackage{subcaption}
\usepackage{pifont}   

\newcommand{\cmark}{\ding{51}} 
\newcommand{\xmark}{\ding{55}} 
\usepackage{hyperref}
\usepackage[skip=5pt]{caption}
\usepackage{tcolorbox}
\usepackage{enumitem}
\title{Normalized Architectures are Natively 4-Bit}

\author{
Maxim Fishman\, ${^\wedge}$\thanks{Equal contribution}  \ \quad
Brian Chmiel\, ${^\wedge}$\footnotemark[1] \ \quad
Ron Banner\ ${^\wedge}$\quad
Daniel Soudry\ ${^\wedge}$$^\circ$\quad
Boris Ginsburg\ ${^\wedge}$\quad
\\[0.2cm]
$^\wedge$ NVIDIA \\
$^\circ$ Department of Electrical and Computer Engineering - Technion, Haifa, Israel
 \\[0.2cm]
\small{\texttt{\{\href{mailto:mfishman@nvidia.com}{mfishman}, \href{mailto:bchmiel@nvidia.com}{bchmiel}, \href{mailto:rbanner@nvidia.com}{rbanner},  \href{mailto:bginsburg@nvidia.com}{bginsburg}\}@nvidia.com}}\\
\small{\texttt{\{\href{mailto:daniel.soudry@gmail.com}{daniel.soudry}\}@gmail.com}}}

\begin{document}

\maketitle

\begin{abstract}
    Training large language models at 4-bit precision is critical for efficiency. We show that nGPT, an architecture that constrains weights and hidden representations to the unit hypersphere, is inherently more robust to low-precision arithmetic. This removes the need for interventions—such as applying random Hadamard transforms and performing per-tensor scaling calculations—to preserve model quality, and it enables stable end-to-end NVFP4 training. We validate this approach on both a 1.2B dense model and hybrid (Mamba-Transformer) MoE models of up to 3B/30B parameters. We trace this robustness to the dot product: while quantization noise remains largely uncorrelated in both standard and normalized architectures, the signal behaves differently. In nGPT, the hypersphere constraint enhances weak positive correlations among the element-wise products, leading to a constructive accumulation of the signal across the hidden dimension while the noise continues to average out. This yields a higher effective signal-to-noise ratio and a flatter loss landscape, with the effect strengthening as the hidden dimension grows, suggesting increasing advantages at scale. A reference implementation is available at \url{https://github.com/anonymous452026/ngpt-nvfp4}.
\end{abstract}

\section{Introduction}

Deploying large transformers at 4-bit precision is becoming essential for efficiency, yet low-bit training remains fragile. In practice, standard transformer architectures often require fixes such as randomized Hadamard transforms (RHT), dynamic per-tensor scaling, or mixed-precision exceptions to maintain model quality \cite{NvidiaNVFP4}---which all require overhead. These interventions can make 4-bit training possible, but they also raise a more fundamental question: 
Can we find an architecture that has an intrinsic 4-bit quantization robustness without requiring these quantization tricks?

In this work, we show that quantization robustness can be architectural. We study \textbf{nGPT} \cite{loshchilov2025ngpt}, a transformer that constrains hidden states and model parameters to the unit hypersphere, and find that this normalization makes the model natively robust to NVFP4 arithmetic. We focus on this low-precision format since it is supported natively by NVIDIA's Blackwell GPUs \cite{NvidiaBlack}. Across diverse architectures, including a 1.2B dense model and hybrid Mamba-Transformer Mixture-of-Experts (MoE) configurations ranging from 400M/600M to 3B/30B, nGPT supports stable end-to-end full NVFP4 training, without RHT, without dynamic per-tensor scaling, and without the divergence typically observed in standard transformer baselines when these operations are omitted.

The central question is why. At first glance, one might expect quantization robustness to come from reducing quantization noise. That is the logic behind most existing approaches: control outliers, improve rounding, rescale tensors, or selectively restore higher precision where noise is too large. Our analysis points to a different mechanism. In nGPT, the advantage does not come from making the quantization noise unusually small. It comes from making the \emph{signal} accumulate more effectively.

To trace this effect, we perform a layer-wise structural analysis on a 3.6B parameters transformer, using NVFP4 quantization. We find that quantization noise remains largely uncorrelated in both standard GPT and nGPT. The key difference appears in the signal. In nGPT, the hypersphere constraint induces weak but consistent positive correlations among the element-wise products inside the dot product. These correlations are tiny at the level of any single coordinate, but they act coherently across thousands of dimensions. As a result, the true dot product grows more constructively, while the quantization noise continues to average out like an incoherent random walk.

This behavior is a direct consequence of the training dynamics under the hypersphere constraint. In a standard GPT, the network can rely on a small number of large, unbounded coordinates to dominate the dot product. In nGPT, this path is blocked: because both activations and weights are normalized, no single element can arbitrarily scale the output. Instead, to produce a large dot product, we observe empirically that the model creates alignment across many coordinates. Specifically, although the marginal distributions of individual weights and activations are similar in both architectures (\cref{fig:pipeline}), the element-wise products in nGPT exhibit a systematic positive correlation across coordinates that is absent in GPT.  That structural bias is weak per coordinate, but overall, across the dimensions it produces a robust signal drift that low-precision noise cannot easily disrupt.

This leads to a different view of low-bit robustness. The main determinant is not the suppression of local quantization error, but the coherence of signal accumulation. Under the hypersphere constraint, the signal scales more like a coordinated sum, while the noise remains essentially incoherent. This creates a higher effective dot-product SNR, and the effect compounds with depth into a substantially flatter loss landscape. In other words, nGPT is structurally robust to low-bit quantization not as a side effect, but as a direct consequence of how normalized dot products accumulate signal.

Our results therefore suggest a shift in emphasis for low-precision model design. Much of the literature treats quantization as a compression problem applied after the architecture is fixed. Our findings suggest that the architecture itself can be made quantization-ready. No post-hoc quantization trick is needed to create the effect we observe. The hyperspherical geometry already biases the model toward a representation in which signal survives 4-bit arithmetic unusually well. As training and inference continue to move toward 4-bit precision and below, this structural property may prove more valuable than increasingly elaborate correction mechanisms layered on top of standard architectures.

Our main contributions are: 
\begin{itemize}
    \item \textbf{Architecture-driven robustness:} We show that nGPT is natively ``quantization-ready'', enabling stable end-to-end NVFP4 training. Unlike standard Transformers, it requires no fixes that introduce overhead like randomized Hadamard transforms (RHT) or dynamic per-tensor scaling to keep model quality.
    \item \textbf{The Signal-Accumulation Mechanism:} We identify that nGPT’s robustness stems from signal coherence rather than noise suppression. By forcing the signal to accumulate constructively across thousands of dimensions, nGPT increases the SNR per layer, allowing the true dot-product signal to outpace incoherent quantization noise.
    \item \textbf{Driven by Training Dynamics:} We demonstrate that this robustness is a direct consequence of training dynamics under the unit hypersphere constraint. By blocking the model from relying on a few large, unbounded coordinates to drive outputs, nGPT forces the optimizer to learn distributed alignments across thousands of dimensions. This structural bias ensures the signal remains coherent even under heavy 4-bit quantization.
     \item \textbf{Empirical Validation across Architectures:} We confirm these advantages across diverse configurations, including a 1.2B dense model and a hybrid Mamba-Transformer Mixture-of-Experts (MoE) architectures of 400m/600m and 3B/30B. In all cases, nGPT maintains stability and achieves lower relative error than standard transformers.
    
\end{itemize}

\section{Related works}

The scaling of large language models into the trillion-parameter regime has driven a massive shift toward low-precision arithmetic, specifically utilizing NVIDIA's NVFP4 format. The first work to successfully demonstrate the viability of fully quantized NVFP4 pretraining was \cite{chmiel2025fp4}. To further reduce the inherent quantization error of NVFP4 formats, subsequent methods such as \cite{fourOverSix} introduced adaptive block scaling, modifying the NVFP4 algorithm to make the distribution of representable values more uniform and reduce error for near-maximal values. The work most directly related to ours--and the current benchmark for stable 4-bit pretraining—is \cite{NvidiaNVFP4}. This work establishes the state-of-the-art recipe for NVFP4 in large scale, demonstrating that to prevent divergence and handle severe block-level outliers, standard Transformers require a complex and overhead-heavy intervention pipeline. Specifically, this "best recipe" relies on Randomized Hadamard Transforms (RHT) to computationally disperse outlier features, dynamic per-tensor scaling, and Stochastic Rounding (SR) for unbiased gradient estimation.
Other concurrent efforts have attempted to mitigate 4-bit quantization errors through differentiable quantization estimators \cite{fp4Msft}. Furthermore, frameworks such as \cite{Quartet,Quartet2} have advanced native FP4 pre-training by constructing unbiased gradient estimators designed specifically to minimize the variance inherent in low-precision backpropagation. 

All of these methods treat quantization robustness as an algorithmic challenge to be patched via post-hoc scaling, rather than as an intrinsic architectural property. In contrast, our investigation focuses on the structural properties of the normalized Transformer (nGPT) \cite{loshchilov2025ngpt}, an architecture that constrains all representations and weights to the unit hypersphere. While nGPT was originally introduced to accelerate training, recent literature has begun to explore the broader geometric implications of this design. For instance, recent work on transferable hypersphere optimization \cite{Ren2026RethinkingLM} has re-evaluated language model scaling under these conditions, demonstrating that hyperspherical constraints facilitate stable optimization, predictable scaling laws, and hyperparameter transferability across different model sizes. However, despite these advancements in scaling, no prior work has analyzed nGPT's behavior in the context of low-precision arithmetic. To the best of our knowledge, we are the first to demonstrate that nGPT's inherent hyperspherical constraints make it natively robust to 4-bit quantization, entirely bypassing the need for overhead-heavy mechanisms.

Our empirical finding that nGPT's normalization leads to a flatter loss landscape with robust signal-to-noise ratios connects to a rich body of theoretical literature on optimization geometry. \cite{weightNorm} showed that weight normalization introduces implicit regularization that guides overparameterized models towards flatter, minimum-norm solutions. Furthermore, additional works reveal that structural constraints inherently bias gradient descent toward flatter minima, as rigorously studied in both linear diagonal \cite{chou2024robust} and homogeneous nonlinear neural networks \cite{homogenous}. Our work provides a critical link between these theoretical observations of hypersphere geometry and the practical demands of ultra-low-precision LLM training. 
 
\section{Analysis of nGPT Quantization Robustness}
\label{sec:AnalysisRobustness}
In the following section, we trace the origins of nGPT robustness using a 12-layer transformer with a hidden dimension of $D=4096$, totaling 3.6B parameters, built on the nanoGPT baseline \cite{Karpathy_nanoGPT_2022}. We first demonstrate that this robustness stems from the signal-to-noise ratio (SNR) in the summation. We then attribute the improved SNR to a stronger underlying signal, and finally we show it is driven by an enhanced correlation between weights and activations in nGPT.
\subsection{Where Does the Robustness Come From?}
\label{sec:robust}
\begin{figure}[t]
    \centering
    \begin{subfigure}[b]{0.48\textwidth}
        \centering
        \includegraphics[width=\textwidth]{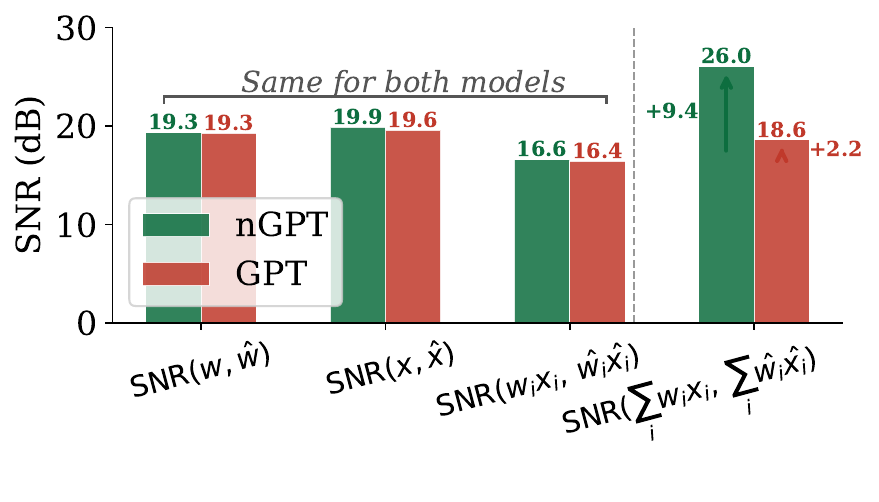}
        \caption{}
        \label{fig:pipeline}
    \end{subfigure}
    \hfill
    \begin{subfigure}[b]{0.48\textwidth}
        \centering
        \includegraphics[width=\textwidth]{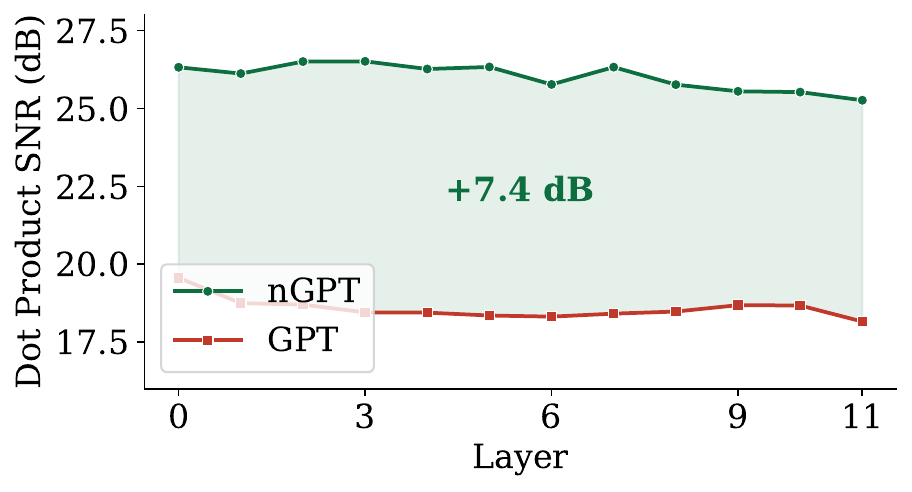} 
        \caption{}
        \label{fig:dotsnr}
    \end{subfigure}
    
    \caption{\textbf{(a):} SNR at each stage of a matrix multiplication under NVFP4, averaged over all 12 layers. The first three stages (individual weights, activations, and their products) are identical for both models. The difference appears only at the final summation step. \textbf{(b):} Dot product SNR under NVFP4 at each layer. nGPT maintains a consistent
$\sim$7\,dB advantage in all layers. Both graphs are based on a 3.6B model, built on the nanoGPT baseline \cite{Karpathy_nanoGPT_2022}. }
\end{figure}

  Every linear layer computes the matrix multiplication $Y = WX$. Each scalar output element $y$ is a dot product $y = \sum_{k=1}^{D} w_k x_k$ of $D{=}4096$ terms. To locate where nGPT's advantage arises, we measure the Signal-to-Noise Ratio (SNR) at each stage of this computation under NVFP4 quantization (\cref{fig:pipeline}). We define the SNR(dB) for a given tensor $T$ and its quantized counterpart $\hat{T}$ as:
\begin{equation}
\text{SNR} = \frac{\|T\|_2^2}{\|T - \hat{T}\|_2^2} \,.
\end{equation}

When dB units are used, we replace SNR with $10 \log_{10}(\text{SNR})$.
\paragraph{Before the sum: no difference.}
The first three groups in Figure~\ref{fig:pipeline} show the SNR of individual
quantities: weights ($w$ vs $\hat{w}$), activations ($x$ vs $\hat{x}$), and
their element-wise products ($w_k x_k$ vs $\hat{w}_k \hat{x}_k$). In all three
cases, nGPT and GPT are nearly identical---both around 19\,dB for elements and
16.5\,dB for products. Normalization does not make individual values easier to quantize.

\paragraph{After the sum: 7\,dB gap.}
The rightmost group shows the SNR of the full dot product
$\sum_k w_k x_k$ vs $\sum_k \hat{w}_k \hat{x}_k$. Here the models diverge sharply:
nGPT achieves 26\,dB; GPT only 18.6\,dB. The improvement from products to dot product---the
\emph{averaging gain}---is +9.4\,dB for nGPT but only +2.2\,dB for GPT.


\paragraph{The advantage is entirely in the summation.}
This is the central finding: normalization does not reduce quantization noise on individual
elements or products---those are identical across architectures. What changes is how
all terms behave when added together. Rather than the noise canceling out, the \emph{signal} in nGPT accumulates constructively during summation, whereas in standard GPT, it does not. Because this coherent signal grows much faster than the accumulated noise, nGPT achieves a fundamentally higher signal-to-noise ratio. The next section explains why.

\paragraph{The gap is consistent across all layers.}
Figure~\ref{fig:dotsnr} shows the dot product SNR at each of the 12 layers individually averaged across all forward matmuls.
nGPT maintains 25--26\,dB uniformly; GPT achieves 18--19\,dB. The advantage is not
concentrated in particular layers---it is a structural property present throughout the network.

\subsection{What drives the SNR in the summation?}
To isolate the structural drivers of dot product SNR, we decompose each matrix multiplication row into a normalized signal and a normalized noise component. Let $s_k = w_k x_k$ denote the $k$-th element-wise product, and let $n_k = \hat{w}_k \hat{x}_k - w_k x_k$ represent the corresponding quantization error, where $\hat{w}$ and $\hat{x}$ are the low-precision quantized values. We define the normalized signal ($z_s$) and normalized noise ($\tilde{z}_n$) by utilizing the signal's random-walk standard deviation as a shared scaling factor:
\begin{equation}\label{eq:zs_zn}
z_s = \frac{\bigl|\sum_{k=1}^{D} s_k\bigr|}{\sqrt{D}\,\sigma_s}, \qquad \tilde{z}_n = \frac{\bigl|\sum_{k=1}^{D} n_k\bigr|}{\sqrt{D}\,\sigma_s},
\end{equation}
where 
$$\sigma_s^2 = {\frac{1}{D-1}\sum_{k=1}^{D}(s_k - \bar{s})^2}, \qquad \bar{s} = \frac{1}{D}\sum_{k=1}^{D} s_k$$
Under this formulation, the dot product SNR factorizes as $\mathrm{SNR}_{\mathrm{dot}} = (z_s / \tilde{z}_n)^2$.

\begin{figure}[t]
\centering
\includegraphics[width=.98\linewidth]{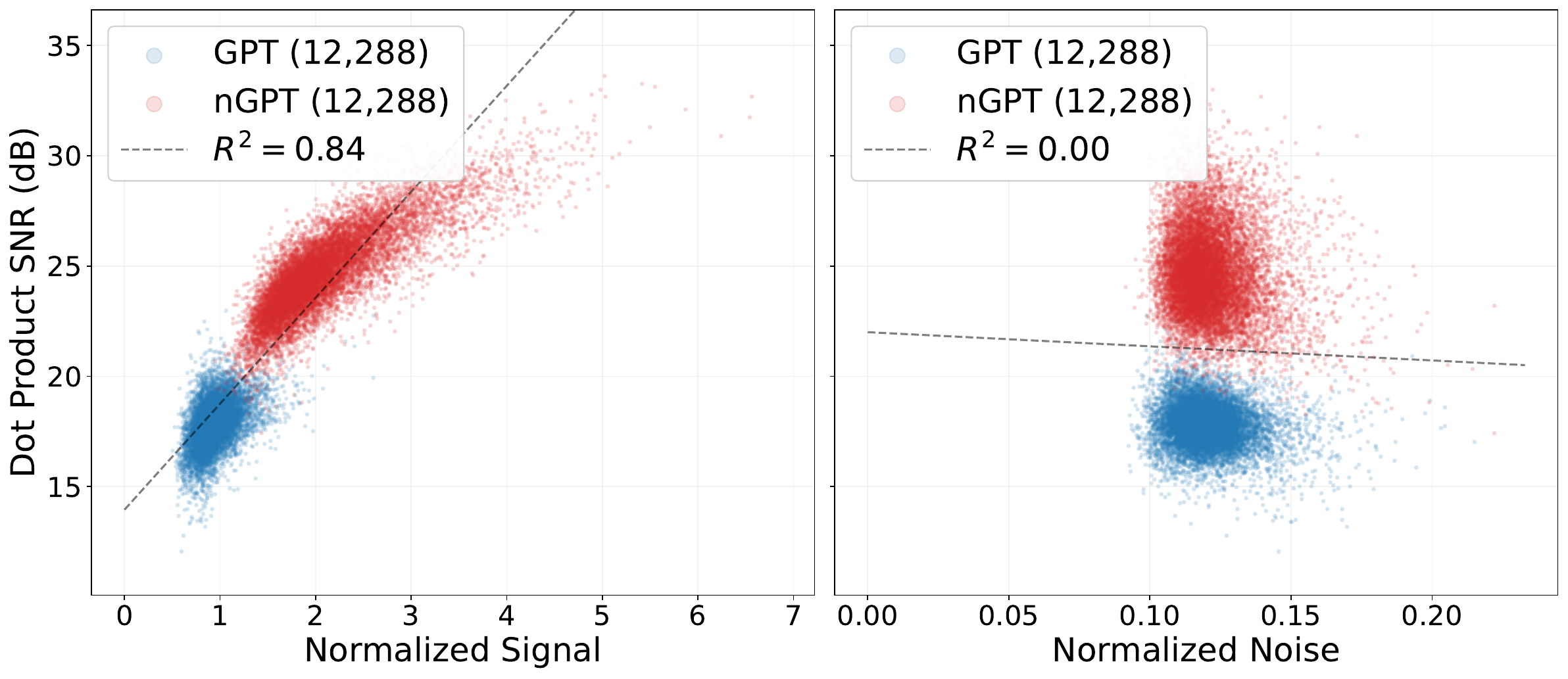}
\caption{\textbf{Normalized signal vs.\ normalized noise as predictors of dot product SNR.}
Evaluated in MLP layers under NVFP4 quantization after training.
\textbf{Left:} The normalized signal $z_s$ demonstrates a strong positive correlation with SNR, cleanly separating the highly coherent representations of nGPT (red) from the baseline GPT (blue).
\textbf{Right:} The normalized noise $\tilde{z}_n$ is nearly identical across both architectures and exhibits no predictive power over the final SNR. This demonstrates that nGPT's SNR advantage originates entirely from coherent signal accumulation rather than structural noise reduction.}
\label{fig:signal_noise_drift}
\end{figure}

Figure~\ref{fig:signal_noise_drift} illustrates the relationship between these decomposed quantities and the empirical dot product SNR evaluated across validation activations. The analysis yields a clear structural distinction: the normalized signal $z_s$ acts as a robust predictor of the overall SNR, establishing a distinct separation between nGPT and standard GPT representations. In stark contrast, the normalized noise $\tilde{z}_n$ remains statistically indistinguishable between the two architectures and demonstrates no predictive capacity for the final SNR. \textbf{nGPT’s superior SNR is driven exclusively by a more robust signal representation rather than a reduction in noise}.

\subsection{Why is the Signal Stronger in nGPT?}
\label{sec:element_correlation}

We now trace the SNR advantage to the structure of the dot-product summation itself. First, we measure statistical dependence between the individual signal and noise terms. Next, we show that nGPT introduces weak but systematic positive correlations in the signal, whereas the noise remains near-uncorrelated in both architectures. Finally, we explain why these correlations strengthen the signal: when positively correlated terms are summed over many dimensions, they accumulate more coherently and produce a larger dot product.

\subsubsection{Measuring Dependence Between Elements in the Dot-Product Sum}
\label{sec:measureCorr}
We begin by directly measuring whether the elements in the dot-product sum behave 
independently or exhibit statistical dependence. If the terms $s_k$ are independent, 
the dot product behaves like a sum of unrelated contributions. However, if they are 
positively correlated, they tend to increase and decrease together, leading to more 
coherent accumulation.

To quantify this, we define an effective pairwise correlation directly from the 
variance decomposition $\mathrm{Var}(\sum_k u_k) = \sum_k \mathrm{Var}(u_k) 
+ 2\sum_{j<k}\mathrm{Cov}(u_j,u_k)$:

\begin{equation}\label{eq:rho_bar}
\bar{\rho}_u \;=\; \frac{1}{D-1}\!\left(
\frac{\mathrm{Var}(\sum_{k=1}^{D} u_k)}
     {\sum_{k=1}^{D} \mathrm{Var}(u_k)} \;-\; 1
\right),
\qquad u \in \{s, n\},
\end{equation}

where $\mathrm{Var}(\,\cdot\,)$ in \cref{eq:rho_bar} denotes variance taken across input samples $x$. 
This quantity is the variance-weighted average of pairwise Pearson correlations 
between elements of the sum, and is equivalent to the standard average correlation when all 
$\mathrm{Var}(u_k)$ are comparable. 


\begin{figure}[t]
    \centering
    \includegraphics[width=\linewidth,height=0.2\textheight]{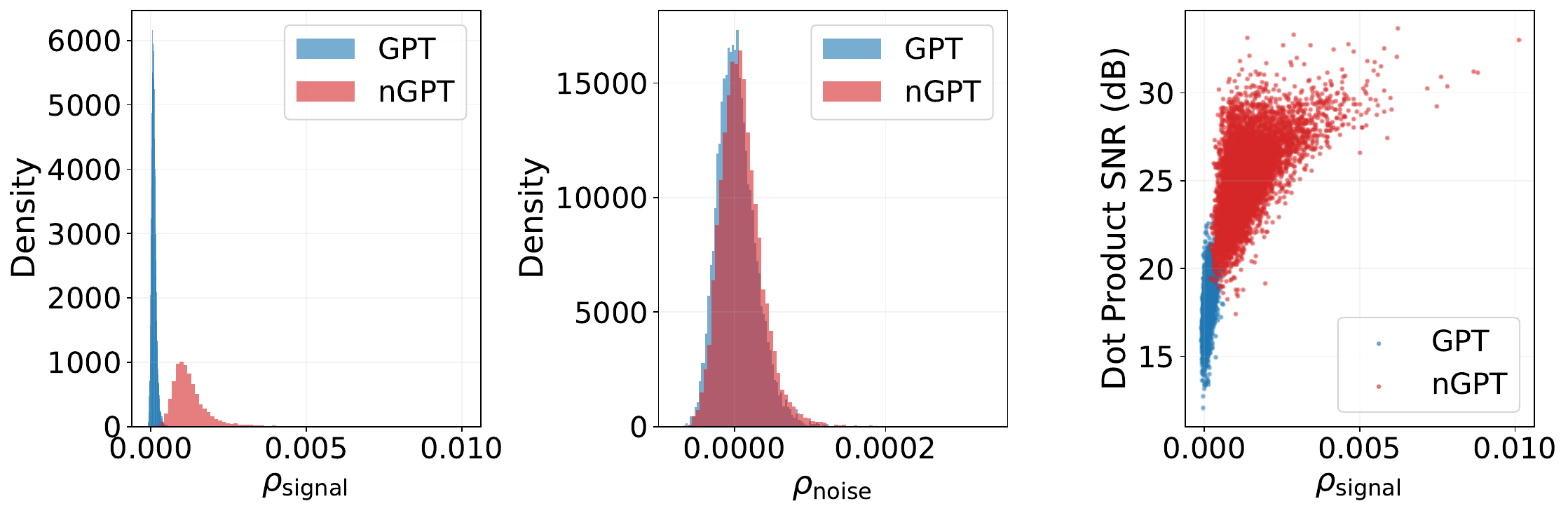}
    \caption{Mean-centered pairwise Pearson correlation of quantized dot-product elements (NVFP4) after training.
    \textbf{Left:} Distribution of signal correlation \(\bar{\rho}_s\); nGPT is shifted toward positive values.
    \textbf{Middle:} Distribution of noise correlation \(\bar{\rho}_n\); both architectures remain near zero.
    \textbf{Right:} \(\bar{\rho}_s\) versus dot-product SNR; higher signal coherence is associated with higher SNR. }
    \label{fig:element_correlation}
\end{figure}

As shown in \cref{fig:element_correlation}, GPT exhibits small signal correlation
(\(\bar{\rho}_s = 9.31 \times 10^{-5}\)),
while nGPT shows a consistently larger correlation
(\(\bar{\rho}_s = 1.32 \times 10^{-3}\), about \(14\times\) larger).
In contrast, noise correlations remain small in both models
(\(\bar{\rho}_n \sim 10^{-6}\)).

\subsubsection{Why correlation makes the signal stronger?}

In \cref{secApp:SNRdot} we show that the SNR of the dot product is: 
\begin{equation}
 \label{SNR}
\mathrm{SNR}
=
\frac{\mathbb{E}[S^2]}{\mathbb{E}[N^2]}
=
\frac{\mathrm{Var}(S)+\mu_S^2}{\mathrm{Var}(N)+\mu_N^2}
\approx
\frac{
D\sigma_s^2 \bigl(1 + (D-1)\rho_s \bigr)+\mu_S^2
}{
D\sigma_n^2+\mu_N^2
}.
\end{equation}
where $\mu_S=\mathbb{E}[S]$ and $\mu_N=\mathbb{E}[N]$. This clearly shows how the positive correlation $\rho_s$ acts as a multiplier on the baseline SNR. 

\paragraph{Summary.} The SNR advantage in nGPT arises from the collective summation of elements rather than individual contributions; this stems not from noise reduction, but from enhanced signal correlation that is inherently stronger in the nGPT architecture. To confirm that increased correlation is a phenomenon driven by nGPT's training dynamics, we trained a one-layer model in \cref{secApp:OneLayerAlign}. The results demonstrate that while both GPT and nGPT achieve comparable training loss, nGPT exhibits higher correlation.

\subsection{Scaling of the SNR Advantage with Width}
\label{subsec:scaling_snr}
We found empirically that $\mu_S^2 / \sigma_s^2 \ll D$ and $\mu_N^2 / \sigma_n^2 \ll D$ and that the ratio $\sigma_s^2 / \sigma_n^2$ is matched across architectures. Therefore, we can use \cref{SNR} to calculate the SNR ratio as follows:


\begin{equation}
\frac{\mathrm{SNR}_{\mathrm{nGPT}}(D)}{\mathrm{SNR}_{\mathrm{GPT}}(D)}
\;\approx\;
\frac{1 + D\,\bar{\rho}_{\mathrm{nGPT}}}{1 + D\,\bar{\rho}_{\mathrm{GPT}}}.
\label{eq:snr_ratio}
\end{equation}

This ratio predicts how much more nGPT benefits from increasing width.

Because $\bar{\rho}_{\mathrm{GPT}} \ll \bar{\rho}_{\mathrm{nGPT}}$, \cref{eq:snr_ratio} exhibits three distinct regimes: small, intermediate and large width.
In \cref{secApp:SNRregimes} we analyze these regimes.

\paragraph{Empirical validation.}
Using the measured values from \cref{sec:measureCorr}  the two transition points are
\[
\frac{1}{\bar{\rho}_{\mathrm{nGPT}}} \approx 755,
\qquad
\frac{1}{\bar{\rho}_{\mathrm{GPT}}} \approx 10{,}738,
\]
and the saturation value is $\bar{\rho}_{\mathrm{nGPT}}/\bar{\rho}_{\mathrm{GPT}} \approx 14.2$.
A key empirical finding is that $\bar{\rho}_s$ is essentially \emph{constant} across all partial summation lengths $k \in [16, 4096]$ for both models.
This means a single measured value of $\bar{\rho}_s$ is sufficient to predict the full width-scaling behavior. Investigating the underlying mechanisms that maintain this constant correlation and why it is inherently higher within the nGPT architecture remains a compelling direction for future research. 

\begin{figure}[t]
    \centering
    \includegraphics[width=.98\linewidth]{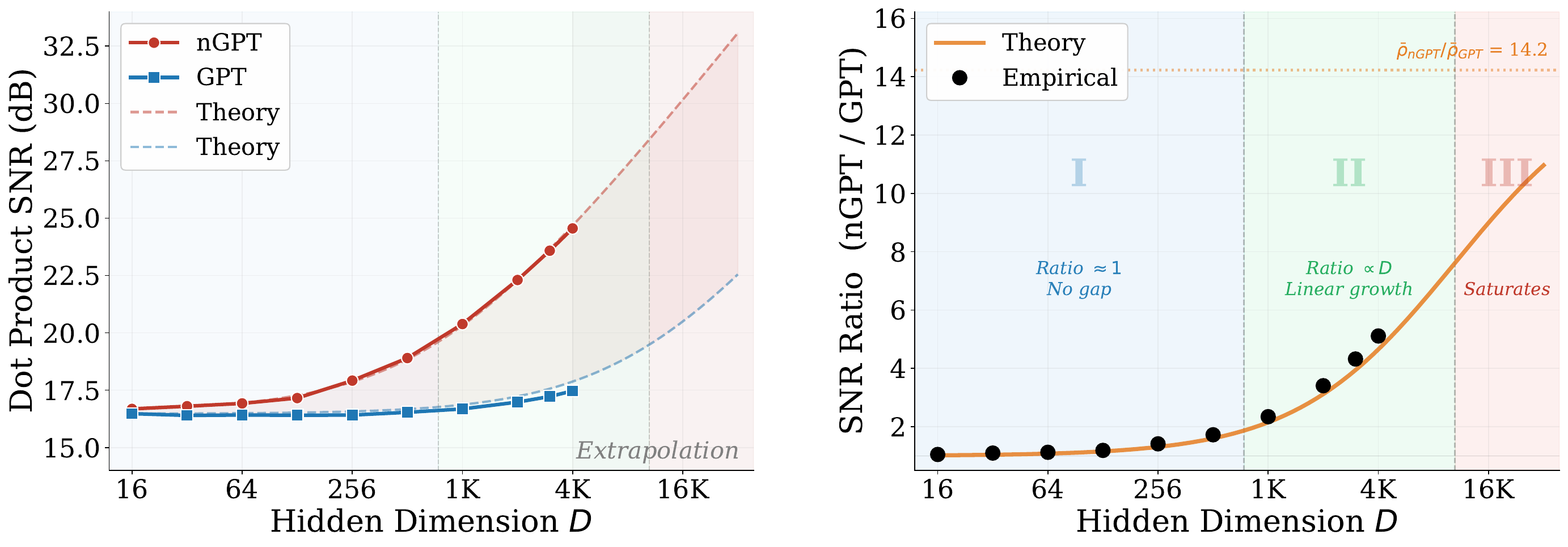}
    \caption{\textbf{Three-regime scaling of nGPT's SNR advantage.}
    \textbf{(Left):} ~Dot-product SNR vs.\ summation length $D$ for both architectures, with theory curves from the measured $\bar{\rho}_s$ (dashed). The theory, which has no free parameters beyond the single measured $\bar{\rho}_s$, closely tracks the data across two orders of magnitude in $D$.
    \textbf{(Right):} ~SNR ratio (nGPT/GPT) vs.\ $D$, showing the three predicted regimes:
    \textbf{I}~(blue, $D \ll 755$): no gap;
    \textbf{II}~(green, $755 \ll D \ll 10738$): linear growth;
    \textbf{III}~(red, $D \gg 10738$): saturation at $\bar{\rho}_{\mathrm{nGPT}}/\bar{\rho}_{\mathrm{GPT}} \approx 14.2$.
    Current models operate in Regime~II; the advantage is expected to continue growing at larger scales. }
    \label{fig:scaling}
\end{figure}

\Cref{fig:scaling}(a) shows the absolute SNR for both models together with the theory curves from \cref{SNR} using the measured $\bar{\rho}_s$. \Cref{fig:scaling}(b) shows the ratio in \cref{eq:snr_ratio}, with the three regimes marked. Current models ($D{=}4096$) sit firmly in Regime~II, where the gap is still growing. At $D{=}16384$ (405B-scale), the theory predicts the ratio will approach~$10\times$.

The practical implication is direct: \emph{larger models benefit more from nGPT}, because positive signal correlation makes SNR grow faster with width.

\section{From Per-Layer SNR to a Flat Loss Landscape}

The averaging gain is measured per layer, but quantization affects all layers simultaneously.
We now show that the per-layer advantage directly predicts a flatter loss landscape for nGPT.

\paragraph{Per-layer: $\sim$ 7 dB improvement in SNR.}
Figure~\ref{fig:mechanism}(a) shows the gain (dB): 
\begin{equation}
    \text{gain(dB)} = \mathrm{SNR}(\sum_i w_ix_i,\, \sum_i \hat{w_i}\hat{x_i})- \mathrm{SNR}(w_ix_i,\, \hat{w_i}\hat{x_i})
    \label{eq:gain}
\end{equation}
 at each of the layers.
At every matrix multiplication, nGPT suppresses quantization noise more effectively than GPT---not because the noise is smaller, but because the signal accumulates faster.

\paragraph{Full network: 3.5$\times$ flatter.}
When all layers are perturbed simultaneously---as happens during quantized training---the per-layer effects compound.
To measure this directly, we add Gaussian noise to all weights at scale $\alpha \cdot \|W\|_F / \sqrt{n}$, where $n$ is the number of entries in $W$, and record the validation loss increase (Figure~\ref{fig:mechanism}b).
GPT's loss degrades $3.5\times$ faster than nGPT's.
Since quantization is a structured perturbation applied to every weight and activation, a flatter landscape means quantization has less impact on the loss.

This connection is direct: a model with higher averaging gain at each layer tolerates more perturbation per layer, and therefore tolerates more total perturbation across the full network.

\section{Experiments}

We now present our empirical evaluation demonstrating nGPT's robustness to NVFP4 quantization. We first analyze the training dynamics and downstream task performance of a 1.2B parameter dense model. Following this, we extend our evaluation to a Hybrid Mamba-Transformer MoE, to confirm that nGPT structural advantages hold across different model topologies. All training was conducted using the Nemotron-Pretraining-Datasets \cite{NemotronData} , AdamW optimizer with $\beta_1=0.9$, $\beta_2=0.95$. All experiments use Blackwell GPUs, which natively support NVFP4 datatype. In \cref{tab:ngpt-comparison} we compare the different training overhead operations, While in nGPT-NVFP4 runs we eliminate the need for Randomized Hadamard Transforms (RHT) and per-tensor scaling, the standard NVFP4 still keeps these operations to avoid divergence \cite{NvidiaNVFP4}. Relative error is defined as $(\text{Loss}_\text{quantized}- \text{Loss}_\text{BF16}) / \text{Loss}_\text{BF16}$ . In \cref{sec:ngptAch} we show the changes required to transform a standard architecture to nGPT.

\subsection{1.2B dense} 
\label{sec:expDense}
\cref{fig:CombinedRelative}(a) shows the relative error of a 1.2B dense model trained on 1T tokens using NVFP4 and nNVFP4. A key advantage of the nGPT architecture is its ability to mitigate the overhead of specific quantization operations, such as RHT and per-tensor-scaling while improving the relative error. As shown in \cref{tab:1_2DenseDown} , the normalized architecture consistently outperforms the standard baseline across various downstream tasks in both BF16 and NVFP4 precisions. In \cref{tab:Dense_model_comparison} we show the hyperparameters used.

\begin{table}[ht]
    \centering
    \small
    \caption{Downstream tasks evaluation of 1.2B dense model. Note that the normalized architecture outperforms the standard architecture over all tasks, both in BF16 and NVFP4 datatype.} 
    \label{tab:1_2DenseDown}
    \vspace{8pt}
    \renewcommand{\arraystretch}{1.4} 
    \setlength{\tabcolsep}{6pt}      
    

    \begin{tabular}{l S[table-format=1.4] S[table-format=2.2] S[table-format=2.2] S[table-format=2.2] S[table-format=2.2] S[table-format=2.2] S[table-format=2.2] }
    \toprule
    \textbf{Config} & {\textbf{Loss} $\downarrow$} & {\textbf{Hell.$\uparrow$ }} & {\textbf{PIQA $\uparrow$}} & {\textbf{Win. $\uparrow$}} & {\shortstack{\textbf{MMLU $\uparrow$} \\ \scriptsize (5-shot)}} & {\shortstack{\textbf{MMLU Pr $\uparrow$} \\ \scriptsize (5-shot)}} & {\shortstack{\textbf{GSM8K $\uparrow$} \\ \scriptsize (8-shot)}} \\ 
    \midrule
        \midrule
        bf16 & 1.47 & 60.54 & 73.72 & 58.17 & 45.23 & 13.98 & 33.66\\
        \rowcolor{gray!10} 
        n bf16 & \bfseries \textbf{1.46} & \bfseries \textbf{61.57} & \textbf{74.27} & \bfseries \textbf{59.51} & \bfseries \textbf{45.50} & \textbf{14.34} & \textbf{36.24} \\
        \hline
        nvfp4 & 1.52 & 58.86 & 73.78 & 58.64 & 42.56 & 13.06 & 29.49 \\
        \rowcolor{gray!10} 
        \parbox[t]{2.5cm}{\textbf{n nvfp4} \\ \tiny (No RHT/No Scale)} & \textbf{1.50} & \textbf{59.93} & \bfseries \textbf{74.37} & 58.64 & \textbf{43.57} & \bfseries \textbf{14.59} & \bfseries \textbf{35.56} \\ 
        \bottomrule
    \end{tabular}
\end{table}

\subsection{Hybrid Mamba-Transformer MoE}
\label{sec:exphybrid}

\paragraph{400m/600m}
In \cref{fig:HybridMoE400}, we compare the training loss and relative error for a Hybrid Mamba-Transformer MoE model (400M/600M) trained over 36B tokens. These results highlight that the benefits of nGPT architecture extend to Hybrid Mamba-Transformer MoE configurations; specifically, it achieves a lower relative error while improving the pipeline through the removal of RHT and per-tensor scaling overhead. In \cref{tab:400exphybrid} we show the hyperparameters used.

\begin{figure}[h]
\centering
\includegraphics[width=\textwidth]{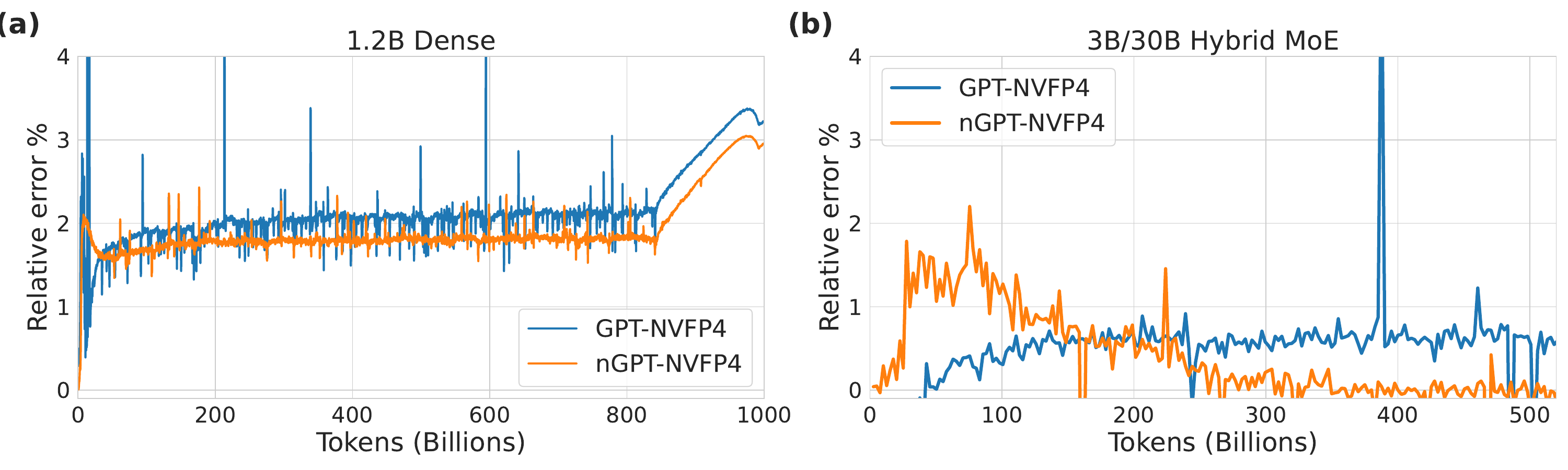}
\caption{\textbf{(a):} 1.2B dense model relative error (1T tokens); nGPT-NVFP4 reduces the loss gap and eliminates RHT/scaling overhead. \textbf{(b):} 3B/30B hybrid MoE relative error; nGPT achieves $\sim$0\% relative error without standard NVFP4 interventions (RHT and per-tensor scaling).}

\label{fig:CombinedRelative}
\end{figure}

\paragraph{3B/30B} In \cref{fig:CombinedRelative}(b), we compare the relative error for a Hybrid Mamba-Transformer MoE model (3B/30B) trained over $\sim$ 500B tokens of a 1T token horizon, achieving $\sim 0 \%$ relative error for nGPT. These results highlight that the benefits of the normalized architecture also extends to large models. Following the methodology in \cite{NvidiaNVFP4}, the NVFP4 experiments maintain high precision for the final 15\% of layers in both GPT and nGPT architectures. In \cref{fig:nanoLoss} we show that both BF16 training runs achieved similar training loss. In \cref{tab:nanoexphybrid} we show the hyperparameters used.  

\subsection{nGPT Learning Rate Robustness Under Quantization}
\label{subsec:lr_sensitivity}

\begin{figure}[t]
    \centering
    \begin{subfigure}[b]{0.48\textwidth}
        \centering
        \includegraphics[width=\textwidth]{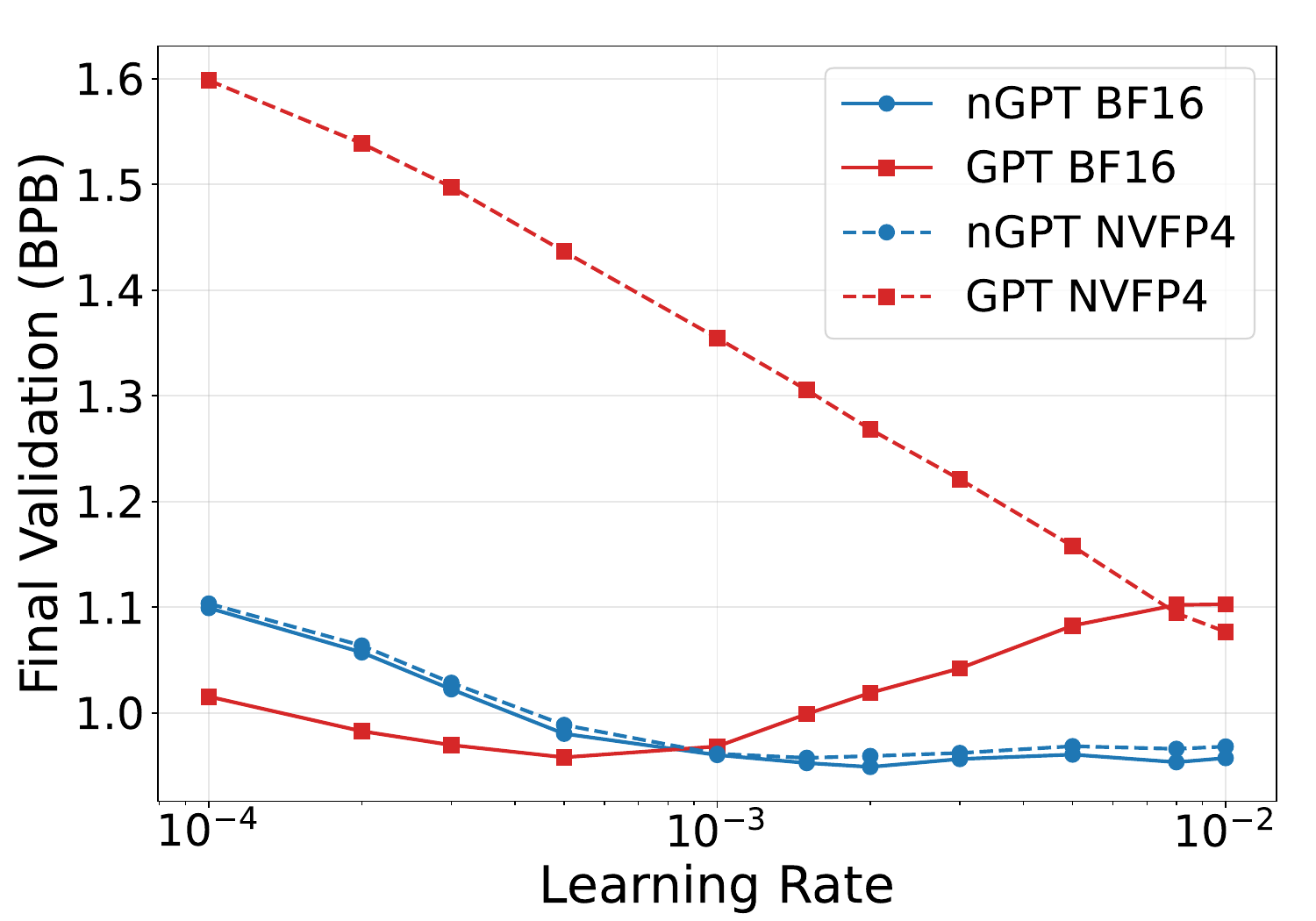}
        \caption{}
\label{fig:lr_sensitivity}
    \end{subfigure}
    \hfill
    \begin{subfigure}[b]{0.48\textwidth}
        \centering
        \includegraphics[width=\textwidth]{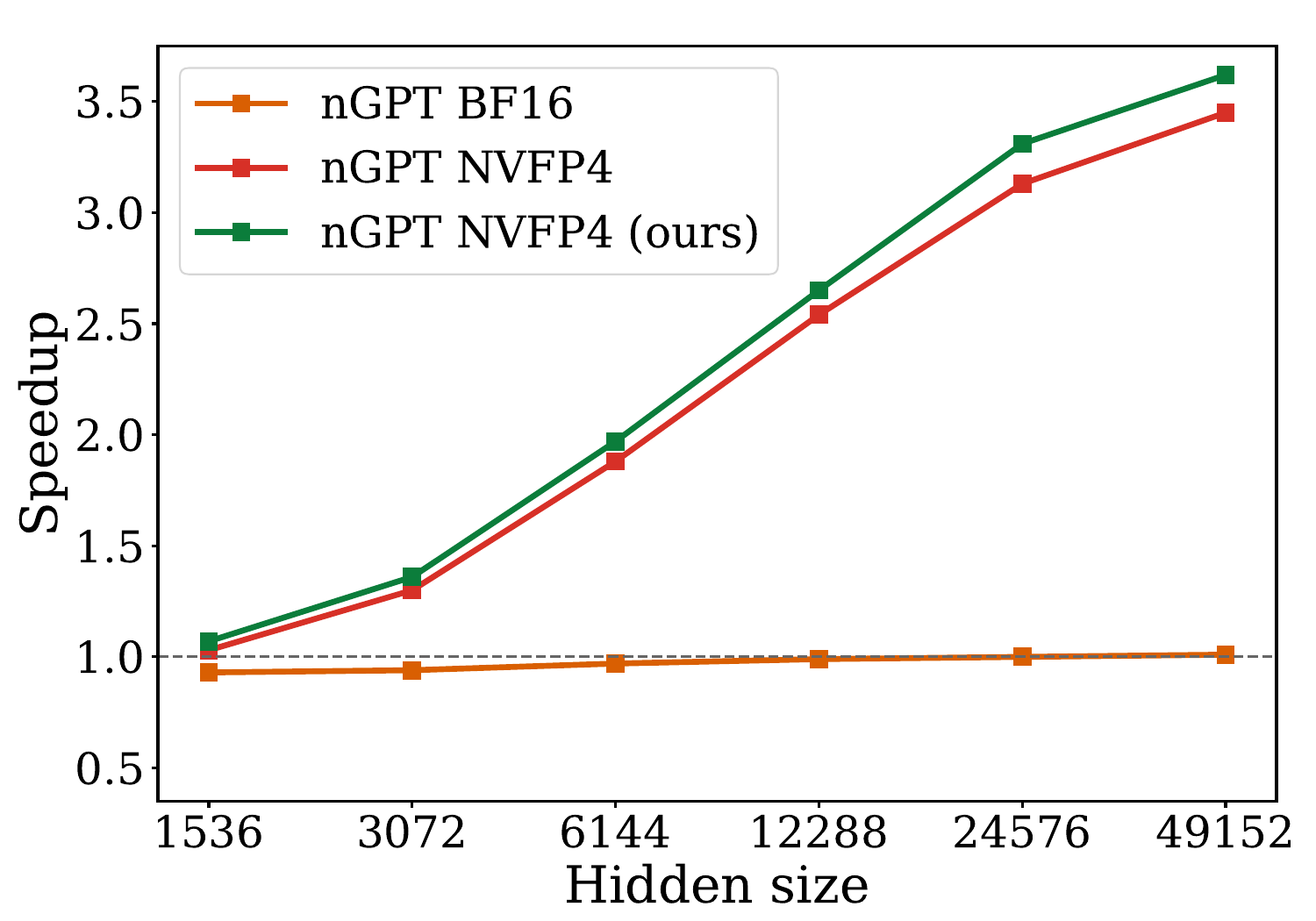} 
        \caption{}
        \label{fig:layer_bench_acceleration}
    \end{subfigure}
    
    \caption{\textbf{(a):} Validation BPB vs. learning rate for nGPT and GPT. nGPT maintains a flat optimum across precisions, while GPT is LR-sensitive under quantization. nGPT’s optimal LR transfers directly from BF16 to NVFP4. \textbf{(b):} One-layer training speedup on a GB200 GPU as a function of hidden size. Speedup is measured relative to the BF16 GPT layer baseline. The nGPT NVFP4 configuration labeled ``ours'' removes both dynamic per-tensor amax scaling computation and RHT.}
\end{figure}

A practical consequence of the higher dot-product SNR (\cref{sec:robust}) is that nGPT should tolerate a wider range of hyperparameters under quantization. To test this, we sweep the learning rate over two orders of magnitude for all four configurations: nGPT and GPT, each in BF16 and NVFP4. We run the same model as in \cref{sec:AnalysisRobustness} and use the final validation bits-per-byte (BPB) as a normalized metric to compare between the runs.  \Cref{fig:lr_sensitivity} reveals two findings:

\paragraph{nGPT is remarkably LR-insensitive.}
In BF16, nGPT achieves nearly identical validation BPB across the entire range. GPT, by contrast, peaks sharply and degrades rapidly at higher rates, reaching a wider spread. This flat minimum is consistent with the flatter loss landscape predicted by the per-layer SNR analysis (\cref{fig:mechanism}b): a model that is less sensitive to weight perturbations is also less sensitive to the effective perturbation introduced by a suboptimal learning rate.

\paragraph{The flat minimum transfers to NVFP4.}
Under NVFP4 quantization, nGPT maintains the same flat profile. The best NVFP4 learning rate for nGPT coincides with the BF16 optimum, with minimal BPB difference at the optimal LR. This means the BF16-optimal hyperparameters can be directly transferred to NVFP4 training without retuning-a significant practical advantage, since hyperparameter sweeps at low precision are expensive. GPT shows the opposite behavior: its NVFP4-optimal LR is $16\times$ larger than the BF16 optimum, requiring a separate sweep.


\subsection{Acceleration}
The same structural property that improves quantization robustness also simplifies the runtime path.  In the standard NVFP4 recipe, activations must be dynamically rescaled and RHT are used to smooth outliers before quantization. nGPT hidden states are already constrained to a bounded hypersphere, so the activation scale can be fixed and the RHT path can be removed. We isolate this effect with the single-layer benchmark on a Blackwell GB200 GPU. Details appear in \cref{secApp:acceleration}.

\paragraph{BF16 nGPT has negligible layer overhead.}
The BF16 nGPT curve remains close to the BF16 GPT baseline across the sweep, indicating that the normalization and interpolation operations themselves do not dominate the layer runtime once the fused kernels are used. The acceleration comes from moving the expensive GEMMs to NVFP4, not from changing the BF16 architecture.

\paragraph{The NVFP4 benefit grows with width.}
\Cref{fig:layer_bench_acceleration} shows that the nGPT NVFP4 path becomes increasingly favorable as the hidden dimension grows. At small widths, fixed costs from Python, autograd, and CUDA launches are still visible; and dominate the runtime. At larger widths, GEMM time dominates and the benefit of Blackwell NVFP4 tensor cores becomes clear: the optimized nGPT NVFP4 path reaches roughly $3.3$--$3.6\times$ speedup over the BF16 GPT baseline at the largest hidden sizes.  Removing both the per-tensor amax reduction and RHT consistently tracks the fastest nGPT NVFP4 configuration in the high-throughput regime, showing that the overhead eliminated by nGPT is not only useful for model quality, but also visible in the measured layer time.

\section{Conclusions}
In this work, we have demonstrated that the robustness of normalized transformers to low-precision arithmetic is not merely a function of quantization algorithms, but a fundamental property of model geometry. 
Our structural analysis reveals that nGPT’s robustness does not stem from a reduction in local quantization noise, but from a superior signal accumulation mechanism. By inducing weak but consistent positive correlations among element-wise products, the architecture ensures that the intended signal accumulates constructively across high-dimensional dot products. The source of this positive correlation in normalized architectures is left for future work.

We identify a unique scaling advantage where nGPT’s signal-to-noise ratio (SNR) grows faster with model width compared to standard architecture. This suggests that as models scale toward larger hidden dimensions, the architectural benefits of normalization become increasingly pronounced.

A critical practical outcome of this structural robustness is nGPT's remarkable insensitivity to LR selection. While standard GPT models exhibit sharp sensitivity to learning rate shifts under quantization—often requiring a large shift in optimal LR when moving from BF16 to NVFP4—nGPT maintains a broad, flat optimum. 

Our empirical evaluations demonstrate that these advantages are consistent across diverse and large-scale architectures. In a 1.2B dense model trained on 1T tokens, nGPT in NVFP4 achieved lower relative error and superior performance across various downstream benchmarks, despite the removal of RHT and per-tensor scaling overhead. These benefits extend to Hybrid Mamba-Transformer MoE configurations, where evaluations of up to 3B/30B variants showed that the normalized architecture maintains significantly lower relative error under NVFP4 quantization than standard baselines. Our results demonstrate that nGPT’s reduced overhead provides a dual benefit: maintaining high model quality while significantly accelerating layer-wise throughput.

\newpage
\bibliography{bibliography}
\bibliographystyle{plainnat}

\newpage

\appendix

\section{nGPT architecture}
\label{sec:ngptAch}
The architectural modifications introduced by nGPT \cite{loshchilov2025ngpt} are outlined below. The highlighted box emphasizes the weight and activation normalization steps, which are fundamental to our low-precision nGPT regime. Notably, in our approach, we normalize all inputs. This contrasts with the official nGPT \href{https://github.com/NVIDIA/ngpt/blob/main/model.py}{implementation}, which omits normalization for the inputs to the out projection GEMM in the attention block and the FFN2 GEMM in the MLP block.

\begin{enumerate}[series=nGPTsteps]
    \item \textbf{Remove} RMSNorm / LayerNorm layers
\end{enumerate}

\begin{tcolorbox}[colframe=purple, colback=white, boxrule=1.5pt, arc=4mm, left=2pt, right=2pt, top=2pt, bottom=2pt]
\begin{enumerate}[resume*=nGPTsteps]
    \item \textbf{Normalize} all weights along embedding dim (after each step)

    \item \textbf{Replace residual updates:}
    \[
    h \leftarrow \text{Norm}(h + \alpha_A(\text{Norm}(\text{ATTN}(h)) - h)), \quad h \leftarrow \text{Norm}(h + \alpha_M(\text{Norm}(\text{MLP}(h)) - h))
    \]
    \[
    \alpha_A, \alpha_M \in R_{\ge 0}^{d_{\text{model}}} \text{ learnable}; \quad \alpha_{\text{init}} = 0.05, \quad \alpha_{\text{scale}} = 1/\sqrt{d_{\text{model}}}
    \]
\end{enumerate}
\end{tcolorbox}

\begin{enumerate}[resume*=nGPTsteps]
    \item \textbf{Attention:} softmax scale $1/\sqrt{d_k} \to \sqrt{d_k}$; normalize \& rescale $q, k$:
    \[
    q \leftarrow \text{Norm}(q) \cdot s_{qk}, \quad k \leftarrow \text{Norm}(k) \cdot s_{qk} \qquad (s_{qk-\text{init}} = 1, \ s_{qk-\text{scale}} = 1/\sqrt{d_{\text{model}}})
    \]

    \item \textbf{MLP rescaling:}
    \[
    u \leftarrow u \cdot s_u, \quad \nu \leftarrow \nu \cdot s_\nu \sqrt{d_{\text{model}}} \qquad (s_{u,v-\text{init}} = 1, \ s_{u,v-\text{scale}} = 1)
    \]

    \item \textbf{Logit rescaling:} $z \leftarrow z \cdot s_z \qquad (s_{z \text{init}} = 1, \ s_{z \text{scale}} = 1/\sqrt{d_{\text{model}}})$

    \item \textbf{Remove} weight decay and LR warmup
\end{enumerate}

\section{Analysis of nGPT quantization robustness -  proofs}
\subsection{SNR of the dot product}
\label{secApp:SNRdot}
To understand the effect of this difference, consider the sum of the signal terms 

$$S = \sum_{k=1}^{D} s_k$$

If the terms are uncorrelated, then the variance of the sum is simply the sum of the variances:$$\mathrm{Var}(S) = \sum_{k=1}^{D} \mathrm{Var}(s_k)$$However, if different terms are positively correlated, the variance increases:

$$\mathrm{Var}(S) = \sum_{k=1}^{D} \mathrm{Var}(s_k) + 2\sum_{j<k} \mathrm{Cov}(s_j, s_k)$$

Assume the signal terms have similar variance $\sigma_s^2$ and an average pairwise correlation $\rho_s$. Then 

$$\mathrm{Var}(S) \approx D\sigma_s^2 \bigl(1 + (D-1)\rho_s \bigr)$$

Because the quantization noise remains largely uncorrelated in both models ($\rho_n \approx 0$), the variance of the noise sum $N = \sum_{k=1}^{D} n_k$ scales strictly linearly: 

$$\mathrm{Var}(N) \approx D\sigma_n^2$$ 

If the sums have non-zero means, let $\mu_S=\mathbb{E}[S]$ and $\mu_N=\mathbb{E}[N]$. Since $\mathbb{E}[S^2]=\mathrm{Var}(S)+\mu_S^2$ and $\mathbb{E}[N^2]=\mathrm{Var}(N)+\mu_N^2$, the SNR of the dot product is:
\begin{equation} \label{eq: snr appendix}
\mathrm{SNR}
=
\frac{\mathbb{E}[S^2]}{\mathbb{E}[N^2]}
=
\frac{\mathrm{Var}(S)+\mu_S^2}{\mathrm{Var}(N)+\mu_N^2}
\approx
\frac{
D\sigma_s^2 \bigl(1 + (D-1)\rho_s \bigr)+\mu_S^2
}{
D\sigma_n^2+\mu_N^2
}.
\end{equation}










\subsection{Scaling SNR regimes}
\label{secApp:SNRregimes}

To derive \cref{eq:snr_ratio} from \cref{eq: snr appendix}, we assume, based on our empirical observations, that $\mu_S^2 / \sigma_s^2 \ll D$ and $\mu_N^2 / \sigma_n^2 \ll D$, and that the ratio $\sigma_s^2 / \sigma_n^2$ are matched across architecture. Based on these assumptions, we can identify three distinct regimes when  $\bar{\rho}_{\mathrm{GPT}} \ll \bar{\rho}_{\mathrm{nGPT}}$:

\paragraph{Regime I: small width, $D \ll 1/\bar{\rho}_{\mathrm{nGPT}}$.}
Both $D\bar{\rho}$ terms are much smaller than~1, so
\begin{equation}
\frac{\mathrm{SNR}_{\mathrm{nGPT}}}{\mathrm{SNR}_{\mathrm{GPT}}}
\;\approx\; 1.
\end{equation}
At small width, neither model accumulates enough correlated signal to create a gap.

\paragraph{Regime II: intermediate width, $1/\bar{\rho}_{\mathrm{nGPT}} \ll D \ll 1/\bar{\rho}_{\mathrm{GPT}}$.}
The numerator is dominated by $D\bar{\rho}_{\mathrm{nGPT}}$, while the denominator remains close to~1:
\begin{equation}
\frac{\mathrm{SNR}_{\mathrm{nGPT}}}{\mathrm{SNR}_{\mathrm{GPT}}}
\;\approx\; D\,\bar{\rho}_{\mathrm{nGPT}}.
\end{equation}
The advantage of nGPT grows \emph{linearly} with width. Every additional dimension contributes to the gap.

\paragraph{Regime III: large width, $D \gg 1/\bar{\rho}_{\mathrm{GPT}}$.}
Both terms are dominated by their correlation contributions, and $D$ cancels:
\begin{equation}
\frac{\mathrm{SNR}_{\mathrm{nGPT}}}{\mathrm{SNR}_{\mathrm{GPT}}}
\;\approx\;
\frac{\bar{\rho}_{\mathrm{nGPT}}}{\bar{\rho}_{\mathrm{GPT}}}.
\end{equation}
The gain saturates at a constant set by the ratio of the two correlations.

\subsection{One layer alignment}
\label{secApp:OneLayerAlign}
\begin{figure}[t]
    \centering
    \begin{subfigure}[b]{0.48\textwidth}
        \centering
        \includegraphics[width=\textwidth]{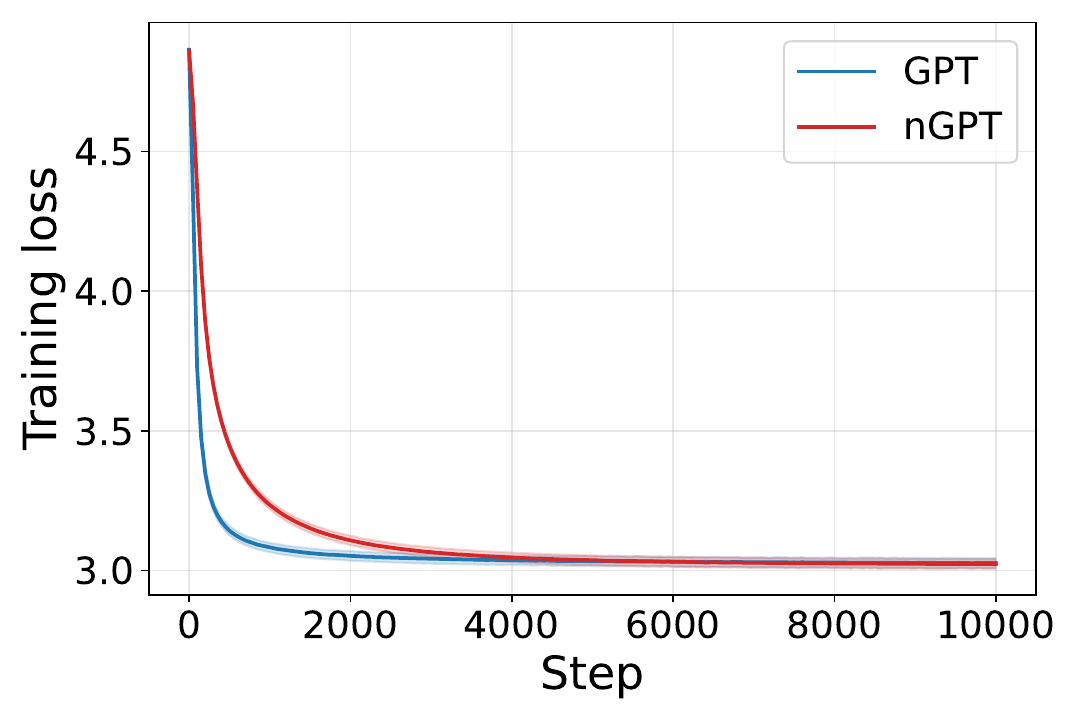}
        \caption{}
        \label{fig:OneLayerTrain}
    \end{subfigure}
    \hfill
    \begin{subfigure}[b]{0.48\textwidth}
        \centering
        \includegraphics[width=\textwidth]{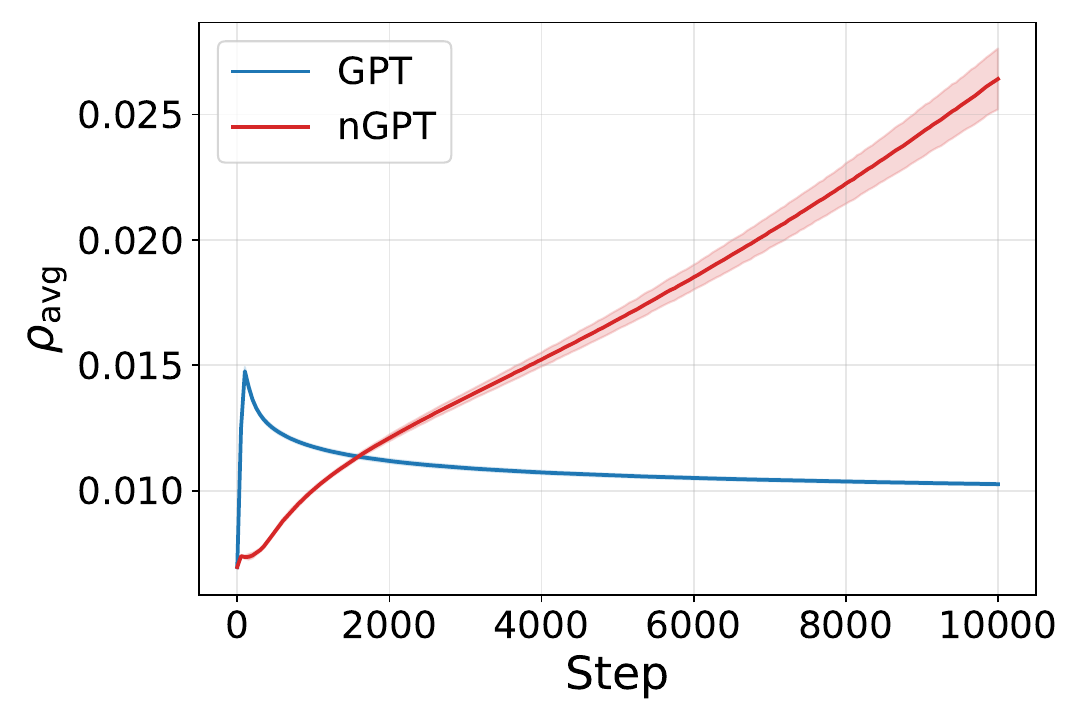} 
        \caption{}
        \label{fig:oneLayerRho}
    \end{subfigure}
    
    \caption{Training loss \textbf{(a)} and average correlation \textbf{(b)} of a one-layer mlp model for GPT and nGPT architectures. While both models achieve similar training loss, the correlation $\rho$ in nGPT is higher.}
\end{figure}

To confirm that nGPT's higher correlation is a reproducible phenomenon rather than model-specific, we trained a single-layer MLP. As illustrated in \cref{fig:OneLayerTrain,fig:oneLayerRho}, both architectures achieve comparable training loss, yet nGPT exhibits significantly higher average correlation, further validating the analysis in \cref{sec:element_correlation}.

\begin{figure}[h]
\centering
\includegraphics[width=\linewidth]{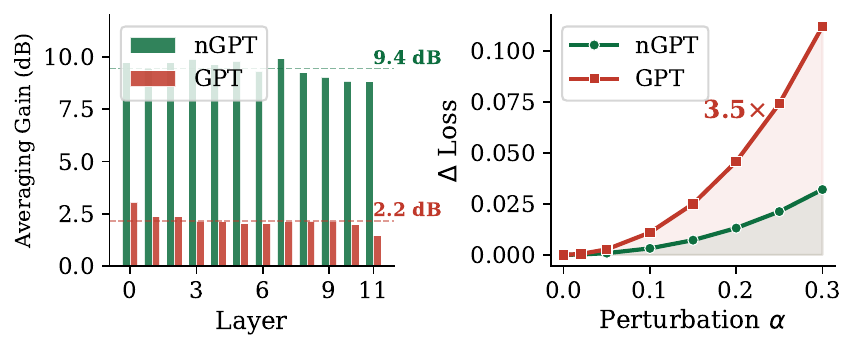}
\caption{\textbf{(Left):} Per-layer gain(dB): nGPT cancels noise better at every matmul. Gain (dB) refers to \cref{eq:gain}. \textbf{(Right):} Loss landscape flatness: perturbing all weights simultaneously, GPT degrades 3.5$\times$ faster. The per-layer advantage compounds into a measurably flatter loss surface.}
\label{fig:mechanism}
\end{figure}

\section{Experiments details}

\begin{table}[h]
\centering
\caption{Comparison of training overhead between nGPT-NVFP4 and GPT-NVFP4}
\label{tab:ngpt-comparison}
\begin{tabular}{@{}lccc@{}}
\toprule
Method & Per-tensor scaling & RHT & SR \\ \midrule
nGPT-NVFP4 & \xmark & \xmark & \cmark \\
GPT-NVFP4  & \cmark & \cmark & \cmark \\ \bottomrule
\end{tabular}
\end{table}

\subsection{1.2B dense}

The hyperparameters for the 1.2B dense model experiments, discussed in \cref{sec:expDense}, are detailed in \cref{tab:Dense_model_comparison}. Notably, the nGPT architecture omits both warmup and weight decay while utilizing a learning rate half that of the standard model. In both 1.2B dense NVFP4 experiments we quantize the GEMMs of all layers. We keep the embeddings, non linear operations and batched-GEMMs in BF16.

\begin{table}[h]
\centering
\caption{Hyperparameters for 1.2B Dense GPT and nGPT models}
\label{tab:Dense_model_comparison}
\begin{tabular}{l|c|c}
\hline
& \textbf{GPT} & \textbf{nGPT} \\ \hline
Number of layers & \multicolumn{2}{c}{20} \\ \hline
Model dimension & \multicolumn{2}{c}{2048} \\ \hline
FFN Dimension & \multicolumn{2}{c}{6144} \\ \hline
Head dimension & \multicolumn{2}{c}{128} \\ \hline
Q heads & \multicolumn{2}{c}{16} \\ \hline
KV heads & \multicolumn{2}{c}{4} \\ \hline
Sequence length & \multicolumn{2}{c}{8192} \\ \hline
Scheduler & \multicolumn{2}{c}{WSD} \\ \hline
Batch size & \multicolumn{2}{c}{768} \\ \hline
Warmup (Samples) & 1M & 0 \\ \hline
lr & $1.2 \times 10^{-3}$ & $0.6 \times 10^{-3}$ \\ \hline
Weight decay & 0.1 & 0 \\ \hline
Init std & 0.0198 & 0.0221 \\ \hline
\end{tabular}
\end{table}

\subsection{Hybrid Mamba-Transformer MoE 400m/600m}
The hyperparameters for the 400m/600m hybrid Mamba-Transformer MOE model experiments, discussed in \cref{sec:exphybrid}, are detailed in \cref{tab:400exphybrid}. Notably, the nGPT architecture omits both warmup and weight decay while utilizing a learning rate half that of the standard model. For both NVFP4 hybrid 400m/600m experiments we quantize the GEMMs of all layers. We keep the embeddings, non linear operations and batched-GEMMs in BF16.

\begin{table}[h]
\centering
\caption{Hyperparameters for 400m/600m hybrid Mamba-Transformer MOE model. The model uses the following hybrid pattern: "MEMEM*EMEMEM" where  "M" refer to Mamba block, "E" to MOE block and "*" to attention block. }
\label{tab:400exphybrid}
\begin{tabular}{l|c|c}
\hline
& \textbf{GPT} & \textbf{nGPT} \\ \hline
Number of layers & \multicolumn{2}{c}{12} \\ \hline
Model dimension & \multicolumn{2}{c}{1024} \\ \hline
FFN Dimension & \multicolumn{2}{c}{1536} \\ \hline
Head dimension & \multicolumn{2}{c}{128} \\ \hline
Q heads & \multicolumn{2}{c}{8} \\ \hline
KV heads & \multicolumn{2}{c}{4} \\ \hline
Sequence length & \multicolumn{2}{c}{8192} \\ \hline
Scheduler & \multicolumn{2}{c}{WSD} \\ \hline
Batch size & \multicolumn{2}{c}{64} \\ \hline
Num experts & \multicolumn{2}{c}{16} \\ \hline
Activated experts & \multicolumn{2}{c}{4} \\ \hline
Mamba num heads & \multicolumn{2}{c}{16} \\ \hline
Mamba head dim & \multicolumn{2}{c}{64} \\ \hline
Warmup (Samples) & 640K & 0 \\ \hline
lr & $1.2 \times 10^{-3}$ & $0.6 \times 10^{-3}$ \\ \hline
Weight decay & 0.1 & 0 \\ \hline
Init std & 0.0198 & 0.031 \\ \hline
\end{tabular}
\end{table}

\begin{figure}[t]
    \centering
    \begin{subfigure}[b]{0.8\textwidth}
        \centering
        \includegraphics[width=\textwidth]{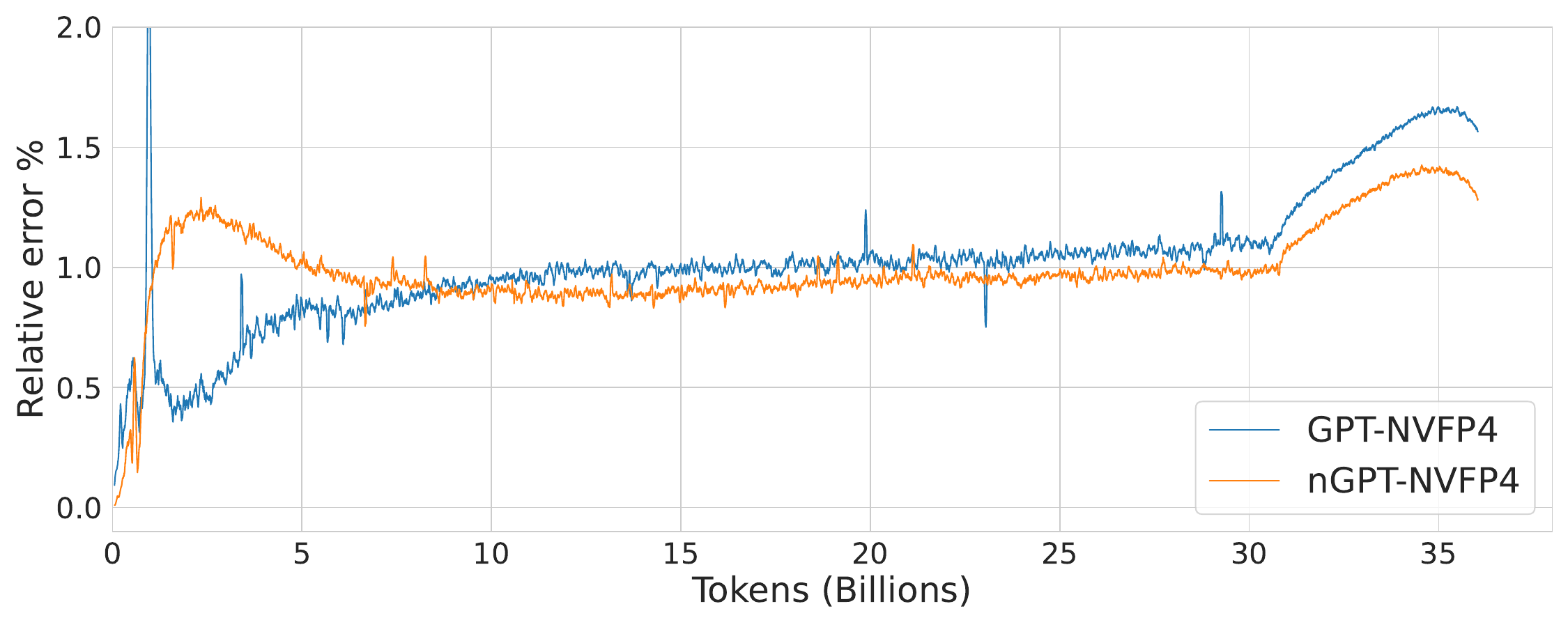}
        \caption{}
        \label{fig:400relative}
    \end{subfigure}
    \hfill
    \begin{subfigure}[b]{0.8\textwidth}
        \centering
        \includegraphics[width=\textwidth]{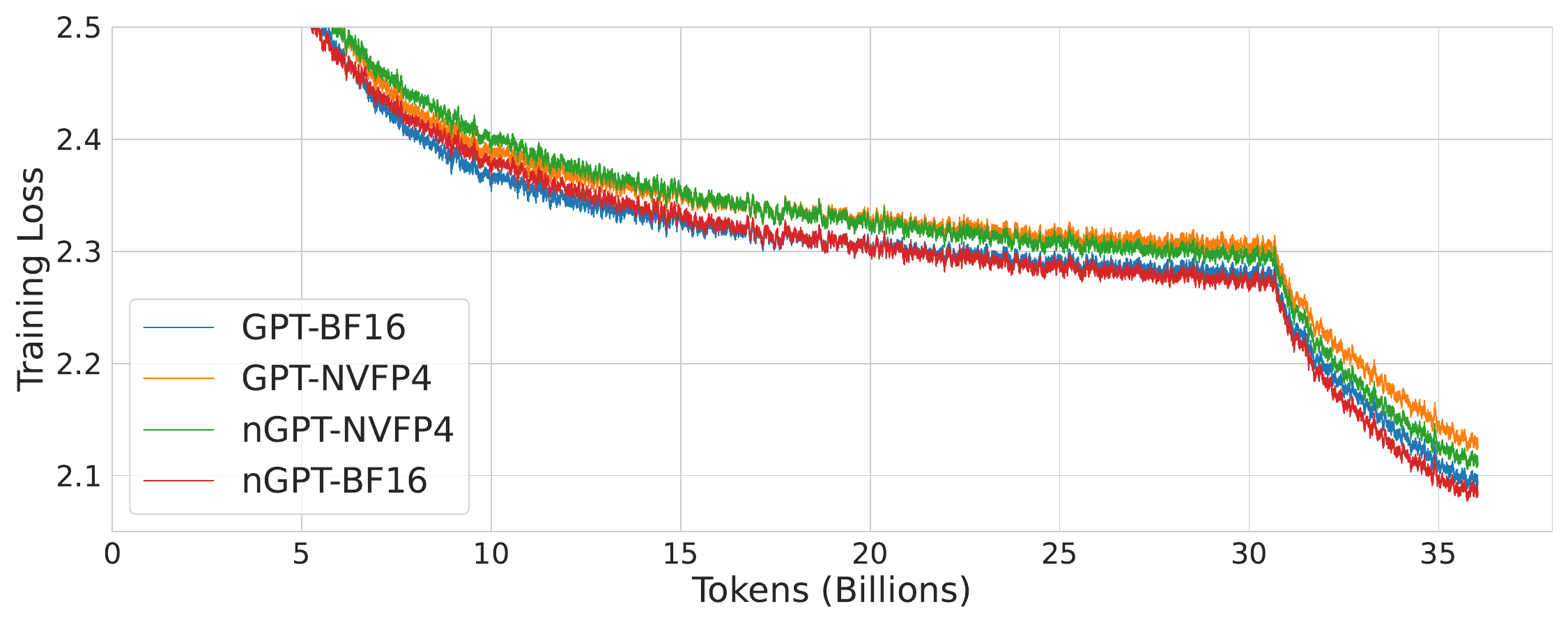} 
        \caption{}
        \label{fig:400_loss}
    \end{subfigure}
    
    \caption{Relative error (a) and training loss (b) of NVFP4 and nNVFP4 with their corresponding Bf16 and nBF16 for hybrid Mamba-Transformer MoE 400m/600m. The normalized architecture achieves lower training loss and lower relative error, while avoiding the use of RHT and per-tensor-scaling. }
    \label{fig:HybridMoE400}
\end{figure}

\subsection{Hybrid Mamba-Transformer MoE 3B/30B}
\begin{figure}[h]
\centering
\includegraphics[width=0.9\textwidth]{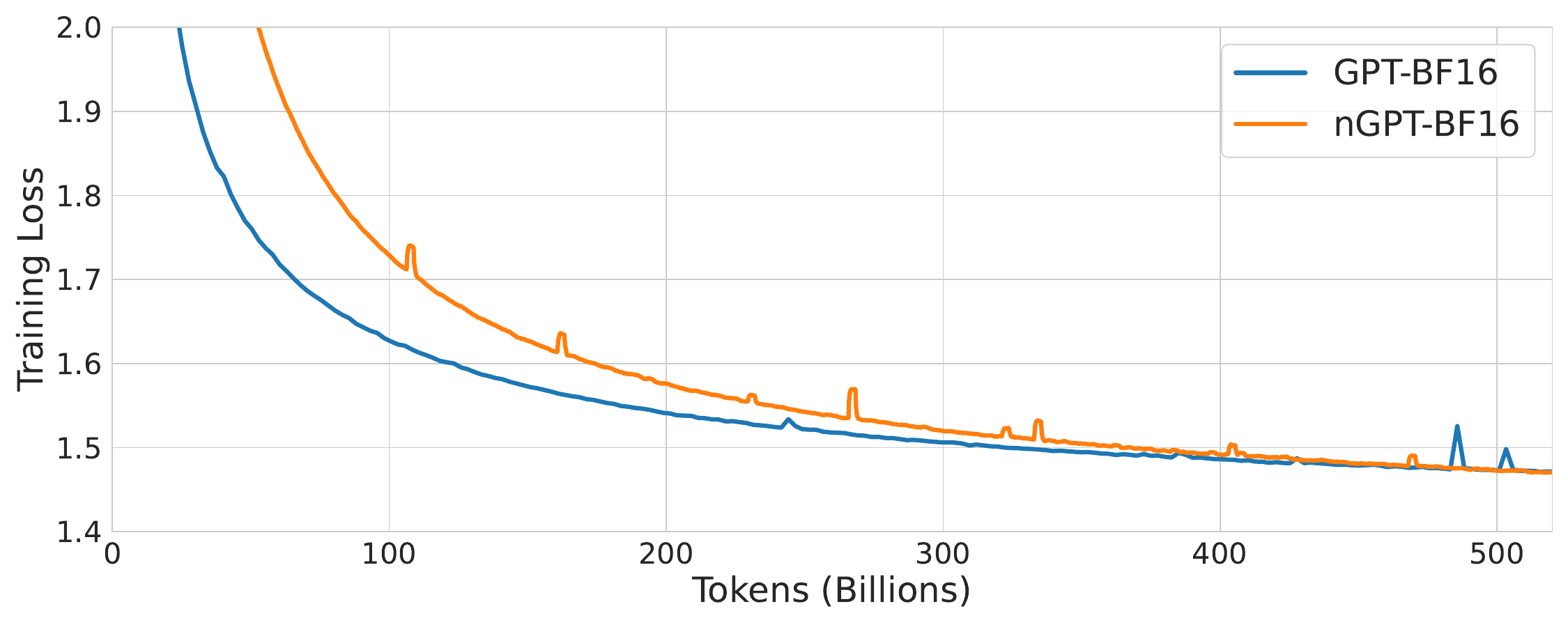}
    \caption{Training loss of nGPT and standard GPT for BF16  datatype for the hybrid Mamba-Transformer 3B/30B model. The architecture is similar to \cite{nanoV3}.  Both training achieved similar training loss. }
    \label{fig:nanoLoss}
\end{figure}

The hyperparameters for the 3B/30B hybrid Mamba-Transformer MoE model experiments, discussed in \cref{sec:exphybrid}, are detailed in \cref{tab:nanoexphybrid}. Notably, the nGPT architecture omits both warmup and weight decay while utilizing a learning rate half that of the standard model. 
In \cref{fig:nanoLoss} we compare the BF16 training of GPT and nGPT, showing they achieve similar training loss. For both hybrid 3B/30B NVFP4 experiments we maintain high precision for the final 15\% of layers in both GPT and nGPT architectures, similar to \cite{NvidiaNVFP4}. In \cref {fig:nanoLoss}

\begin{table}[h]
\centering
\caption{Hyperparameters for 3B/30B hybrid Mamba-Transformer MoE model. The model uses the following hybrid pattern: "MEMEM*EMEMEM*EMEMEM*EMEMEM*EMEMEM*EMEMEMEM*EMEMEMEME" where  "M" refer to Mamba block, "E" to MOE block and "*" to attention block.}
\label{tab:nanoexphybrid}
\begin{tabular}{l|c|c}
\hline
& \textbf{GPT} & \textbf{nGPT} \\ \hline
Number of layers & \multicolumn{2}{c}{52} \\ \hline
Model dimension & \multicolumn{2}{c}{2688} \\ \hline
FFN Dimension & \multicolumn{2}{c}{1856} \\ \hline
Head dimension & \multicolumn{2}{c}{128} \\ \hline
Q heads & \multicolumn{2}{c}{32} \\ \hline
KV heads & \multicolumn{2}{c}{2} \\ \hline
Sequence length & \multicolumn{2}{c}{8192} \\ \hline
Scheduler & \multicolumn{2}{c}{WSD} \\ \hline
Batch size & \multicolumn{2}{c}{3072} \\ \hline
Num experts & \multicolumn{2}{c}{128} \\ \hline
Activated experts & \multicolumn{2}{c}{6} \\ \hline
Mamba num heads & \multicolumn{2}{c}{64} \\ \hline
Mamba head dim & \multicolumn{2}{c}{64} \\ \hline
Warmup (Samples) & 1M & 0 \\ \hline
lr & $1 \times 10^{-3}$ & $0.5 \times 10^{-3}$ \\ \hline
Weight decay & 0.1 & 0 \\ \hline
Init std & 0.0173 & 0.0192 \\ \hline
\end{tabular}
\end{table}

\subsection{Acceleration}
\label{secApp:acceleration}

The experiments in \cref{fig:layer_bench_acceleration} include a single transformer layer benchmark with the same transformer-layer structure as the training runs with RMSNorm fused into GEMMs, SwiGLU MLPs, no linear bias, and input gradients enabled so that backward includes the activation-gradient GEMMs. Forward and backward are timed in the same iteration, with an explicit output gradient and with gradient cleanup excluded from the measured time.  We sweep over multiple hidden sizes ($D$), use FFN size $3D$, set the number of attention heads to $D/128$, disable GQA, and run with MBS one. Each point is averaged over 10 independent repetitions of 20 timed iterations.

\section{Broader Impact}
Accelerating LLM runtime is pivotal as AI becomes increasingly embedded in daily life. By making 4-bit training an intrinsic architectural feature rather than a complex patch, this work improves efficiency, reduces memory demands, and lowers the carbon footprint of large-scale deployment. These advancements democratize AI by enabling high-performance models to run on consumer hardware, fostering broader innovation. However, while increased accessibility empowers diverse users, it also underscores the necessity for responsible oversight and ethical development to mitigate potential misuse as these powerful tools become more widely available.

\section{Limitations}

While nGPT shows clear architectural advantages, our analysis primarily uses smaller models ($D=4096$) to isolate the SNR mechanism. Additionally, while the 3B/30B MoE results demonstrate stability, the 500B token training duration is a relatively short horizon compared to the 20T+ tokens used for SOTA production models; it is possible that performance gaps could evolve over longer training periods or at even greater scales.

\newpage



\end{document}